\documentclass[pmlr,twocolumn,10pt]{jmlr}

\mlhtrack{proceedings}

\newif\iffinal
\finaltrue  % preprint: named authors, no line numbers

\jmlrproceedings{}{Preprint}
\jmlrworkshop{}

\usepackage{booktabs}
\usepackage{graphicx}
\usepackage{tikz}
\usetikzlibrary{positioning}
\usepackage{siunitx}
\usepackage{url}
\usepackage{orcidlink}
\usepackage[switch]{lineno}
\iffinal\else\linenumbers\fi

\usepackage{caption}
\title[Action-Level Reliability of Clinical LLM Agents]{Same Patient, Different
Order: Action-Level Reliability of Clinical LLM Agents Under Repeated Runs}

\author[Bellibatlu, Singh, Han and Zhang]{%
  Rohith Reddy Bellibatlu~\orcidlink{0009-0003-6083-0364}\textsuperscript{*,1},
  Manpreet Singh~\orcidlink{0000-0003-2368-2377}\textsuperscript{2},
  Zhoutian Han\textsuperscript{3}, and
  Wenbin Zhang~\orcidlink{0000-0003-3024-5415}\textsuperscript{1}\\[0.75em]
  {\small\textsuperscript{1}Florida International University, Miami, FL, USA\\
   \textsuperscript{2}Boston University, Boston, MA, USA\\
   \textsuperscript{3}Stevens Institute of Technology, Hoboken, NJ, USA}\\[0.45em]
  {\small\textsuperscript{*}Corresponding author:
   \href{mailto:rohithreddybc@gmail.com}{\textcolor{black}{rohithreddybc@gmail.com}}}%
}
\jmlrauthors{Bellibatlu, Singh, Han and Zhang}

\begin{document}

\maketitle

% HAN 2026-09-09 (verbatim): i think you should tell what's your key idea in
% abstract, not only just report you released some reproductive codebases. For
% readers and reviewers if they can't find something interesting at a glance, they
% are likely to move to other articles.
% HAN-STATUS: done. His own rewrite of the abstract is used, closing on what the
% findings motivate. One clause was added back naming the evidence for the
% single-run claim, which was verified after his revision.
\begin{abstract}
A clinical agent benchmark can report the same verdict on identical inputs while the
agent files a materially different order on each run. Such agents order tests,
request medications and place referrals, yet benchmarks typically score one run per
task and rarely ask whether identical inputs produce identical actions;
MedAgentBench, the benchmark we use, scores a single attempt and says so. To measure
this gap we introduce \emph{same-input rerun}, which replays a task with every input
held fixed and compares the orders rather than the score, with six reliability
metrics, and apply it to 1000 MedAgentBench runs across 50 tasks from its five
write-capable families, two open-weight models below ten billion parameters
quantised to four bits, and two temperatures. The study establishes that
action-level divergence exists and can pass unrecorded by the score, not that any
rate generalises. Under the 8B model at
temperature 0.7, all 43 ordering groups emit a different set of orders across five
identical runs, 26 emit the order on some runs and not others, and 28 record a different
coded value, dose or analyte. In 22 of those 43 the benchmark reports the same
failing verdict for materially different behaviour, as it does for all 10 divergent
groups of the 4B model at 0.7. Orders also reach different endpoints across runs,
one of which the record server rejects while the agent is told it succeeded. These
findings motivate repeated-run evaluation, action-level stability reporting and
execution-faithful environment feedback in clinical-agent benchmarks.
\end{abstract}


\paragraph{Data and Code Availability.}
The harness, the analysis code, the environment reproduction script and all 1250
run records, the grid and the eight-bit arm alike, are at
\url{https://github.com/rohithreddybc/clinical-agent-action-reliability}. An
anonymised read-only mirror of the same commit is served at
\url{https://anonymous.4open.science/r/clinical-agent-action-reliability}; it
carries the identical file tree, so either address reaches the artefact if the
other becomes unavailable. The records include full transcripts, so every number here is
recomputable by the supplied scripts. The patient records are MedAgentBench's own,
described by its authors as real cases de-identified at source, timestamps jittered
and identifiers regenerated \citep{jiang2025medagentbench}. We republish neither the record store nor
any patient data beyond what the environment returned into a transcript, which a run
record must keep if the agent's behaviour is to be audited at all.

\paragraph{Institutional Review Board (IRB).}
We collected no data. The records are those distributed with MedAgentBench, which
its authors report as de-identified at source, timestamps jittered and identifiers
regenerated \citep{jiang2025medagentbench}. We received them in that form, attempted
no re-identification and computed locally, so the work is secondary analysis of an
existing de-identified dataset rather than human subjects research, and no IRB
review was sought.

\paragraph{Use of AI Tools.}
The research question, the study design, the six metric definitions of
Section~\ref{sec:metrics} and every claim made here are the authors', who take full
responsibility for the content. Working to those specifications, an AI assistant
implemented much of the harness and the analysis code. That work is verified rather
than trusted: every number is regenerated from the run records by the released
scripts, an automated gate re-derives each table and fails if the manuscript
disagrees, and the authors re-derived the metric definitions against the FHIR
payloads and confirmed each discordance count and $p$ by hand.

\begin{figure}[t]
\centering
% Colours match Figure 2 so the two read as one visual system. Every colour cue
% is backed by a shape or a word, so nothing is lost in greyscale printing.
\definecolor{okgreen}{HTML}{2B8A3E}
\definecolor{warnred}{HTML}{C2255C}
\definecolor{stepblue}{HTML}{4C6EF5}
\begin{tikzpicture}[
  box/.style={draw=black!55, rounded corners=2pt, align=center, inner sep=3pt,
              font=\footnotesize, fill=black!3},
  run/.style={draw=stepblue!70, rounded corners=1.5pt, align=center,
              inner sep=1.8pt, minimum width=7.5mm, font=\scriptsize,
              fill=stepblue!7},
  lay/.style={align=left, font=\scriptsize},
  emit/.style={align=center, font=\scriptsize, inner sep=0.5pt},
  key/.style={align=center, font=\tiny, text=black!65}]

\node[box, minimum width=34mm] (task)
  {one task, one patient, inputs held fixed};

\node[run, below=5mm of task] (r3) {run 3};
\node[run, left=1.6mm of r3] (r2) {run 2};
\node[run, left=1.6mm of r2] (r1) {run 1};
\node[run, right=1.6mm of r3] (r4) {run 4};
\node[run, right=1.6mm of r4] (r5) {run 5};

\foreach \r in {r1,r2,r3,r4,r5} {\draw[->,gray] (task) -- (\r);}

% What each run emitted. These are the three outcomes the paper actually
% observes, so the reader sees the failure modes before the metric names.
\node[emit, below=1.6mm of r1, text=okgreen] (o1) {$\bullet$};
\node[emit, below=1.6mm of r2, text=okgreen] (o2) {$\bullet$};
\node[emit, below=1.6mm of r3, text=warnred] (o3) {$\circ$};
\node[emit, below=1.6mm of r4, text=okgreen] (o4) {$\bullet$};
\node[emit, below=1.6mm of r5, text=warnred] (o5) {\textbf{--}};

% Kept to two short lines: as one line the legend ran 39pt past the column.
\node[key, below=1.2mm of o3] (key)
  {\textcolor{okgreen}{$\bullet$} order to the typed endpoint \quad
   \textcolor{warnred}{$\circ$} same order, other endpoint\\
   \textcolor{warnred}{\textbf{--}} no order at all};

\node[box, below=1.4mm of key, minimum width=50mm]
  (cmp) {compare the five runs against each other};
\draw[->,gray] (key) -- (cmp);

\node[lay, below=2mm of cmp] (layers)
  {\textbf{orders}: same payload, same endpoint, emitted at all?\\
   \textbf{trajectory}: at which turn do the runs first differ?\\
   \textbf{verdict}: does the benchmark's pass/fail change?};
\draw[->,gray] (cmp) -- (layers);
\end{tikzpicture}
\caption{The rerun protocol, with an illustrative outcome. Inputs are held fixed, so
any difference between runs belongs to the agent, and the three comparison layers
come apart in practice.}
\label{fig:protocol}
\end{figure}
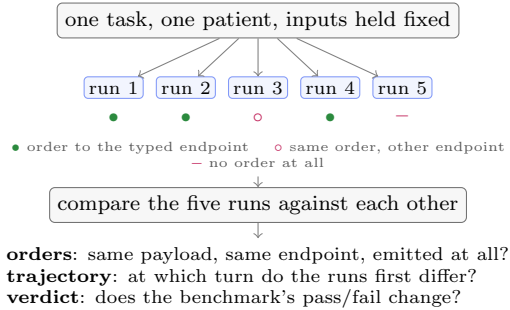

% HAN 2026-09-09 (verbatim): For the Introduction part, it should be looked like
% enhenced version of Abstract part. At the end of Introduction you can express your
% contributions more detailedly. Not only a protocol a codebase or something like
% that but the methodology of clinic agent benchmark designment.
% HAN-STATUS: done. The contributions paragraph now states an evaluation unit, a
% measurement layer and a standard of evidence, rather than listing artefacts.
\section{Introduction}

Clinical language-model agents no longer only answer questions. They retrieve
patient data, order tests, draft prescriptions, and issue write requests against
the record system \citep{jiang2025medagentbench,ferber2026mira,liu2026physicianbench},
and benchmarks have followed them into interactive environments
\citep{schmidgall2024agentclinic}. The reporting convention has not followed:
these benchmarks report accuracy, and mostly from a single run per task. For question
answering the cost is precision, and it is documented, since repeated evaluation of
the same model on the same items produces different scores and rankings built from
single runs invert under replication
\citep{atil2024nondeterminism,madaan2024variance,alvarado2025repetitions}.

For an agent that writes orders, something else goes missing. Accuracy records
whether the agent was right, not whether it was consistent, and only one of those
is reported. The same objection has been raised for static clinical decision support, cleared on
one aggregate figure that says nothing about reliability under shift
\citep{bellibatlu2026rised}. Agents inherit that problem and add one of their own,
because what they produce is an action.

This paper measures consistency. We take MedAgentBench
\citep{jiang2025medagentbench}, a clinical agent benchmark built on a
FHIR-compliant record environment, and replay each task five times with every input
held fixed. Instead of asking whether the verdict changed, we compare what the agent
did: which resource it posted, where, and whether it posted at all. What we propose follows directly: that clinical agent benchmarks
rerun each task and report how often the action changes, that they report an
action-level stability measure beside accuracy, and that their environments answer
as the record server would rather than confirming whatever parses.

Doing this is harder than rerunning a benchmark more times, and three difficulties
shape everything that follows. \textbf{(i) Deciding when two orders are the same
order.} Runs that differ textually can be clinically identical, since timestamps, a
workflow status field and the order of query parameters all vary without the order
varying. Comparing raw transcripts inflates divergence and comparing only the
verdict erases it, so what is needed is a rule for when two requests are the same
order, and every choice inside that rule moves the number. \textbf{(ii) Judging an action the environment never carries out.}
MedAgentBench parses the agent's write request, replies that it succeeded, and does
not write, so nothing inside a run separates an order the record server would have
created from one it would have refused; settling that means leaving the benchmark
loop and putting the requests to the server itself. \textbf{(iii) Reporting a rate
that still means something outside the study that produced it.} A rate measured over
five reruns is not the rate over three, groups that never order have to stay out of
the denominator, and the groups that remain are not independent.
Sections~\ref{sec:metrics}, \ref{sec:endpoint} and \ref{sec:results} take these in
turn.

Three findings follow. Action-level divergence appears in every arm we measured, and
order-set divergence and omission both rise with decoding temperature in the two
models we tested. The mechanism is often mechanical: an order is routed to an
endpoint that would reject it, or is lost after a malformed payload and a broken
retry. Not always, though. In the sampling arm of the larger model 28 of 43 ordering
groups record a different coded value, dose or analyte across identical runs, which
no routing or parsing account explains. And among the tasks that order, the verdict
changes less often than the orders do in every arm.

Taken together these amount to a way of designing and reporting clinical agent
benchmarks, not only the code that implements it here.
\textbf{A protocol.} Same-input rerun makes the unit of evaluation a task group of
identical reruns rather than one attempt, with the ordering groups as the
denominator any action-level rate is quoted over (Section~\ref{sec:metrics}).
\textbf{A measurement layer beneath the verdict.} Six reliability metrics, four
reading the order itself and two the run that produced it, each computable from
transcripts a benchmark already keeps, and one reported in a form that does not
depend on how many reruns were bought. \textbf{Evidence to match.} Interval
estimates and exact paired tests on matched groups, a cluster bootstrap over
families (Section~\ref{sec:results}), and one metric checked against the record
server rather than against our own transcripts (Section~\ref{sec:endpoint}).

\section{Background}
\label{sec:background}

\paragraph{Variance in language model evaluation.}
\citet{atil2024nondeterminism} report accuracy variation across runs of models
configured to be deterministic, which \citet{yuan2025numerical} trace partly to
floating-point non-associativity under changing batch sizes.
\citet{madaan2024variance} say when a difference is meaningful,
\citet{alvarado2025repetitions} find single-run leaderboards invert pairwise
rankings often enough to justify repetition, and \citet{potamitis2025reasonbench}
characterise the variance structure of reasoning systems, while
\citet{mustahsan2025stochasticity} apply intraclass correlation to agentic
evaluation on general-purpose benchmarks.

\paragraph{Repeatability in medicine.}
Clinical work on this question is more developed than the agent literature.
Repeatability has been measured on board-style examinations
\citep{krishna2024radiology}, on unstructured clinical notes
\citep{shah2024consistency}, and on licensing and rare-disease items, where
\citet{shyr2025repeatability} separate repeatability from reproducibility and
\citet{gurses2026consistency} report that single-trial accuracy hides both
volatility and stable systematic error. Closest to a decision with consequences,
\citet{franc2024triage} measure both for a commercial model assigning emergency
triage levels.

% HAN 2026-09-09 (verbatim): intruduce gap first then give your ideas follow behind.
% Next are your verification procedures, so the readers won't forget what problems
% you are trying to address and your core contributions leave readers deeper
% impression. The core pricinple is that reapting your contributions again and again
% through out the article.
% HAN-STATUS: partly done. The proposal now appears in the introduction directly
% after the gap, and again in the abstract's closing sentence. A full reordering of
% the sections has not been attempted.
\paragraph{What is missing.}
Every one of those medical studies scores a single-turn output: an examination
answer, a triage level, a summary of a note. The gap is not merely that none reruns
an agent benchmark; it is that agreement rates and intraclass correlations describe
a scalar or a string, not an order that went to the wrong endpoint or one that was
never issued at all. Several lines of work come closer. $\mathrm{pass}^k$, the
probability that all $k$ trials of a task succeed \citep{yao2024taubench}, scores end
states; \citet{gupta2026reliabilitybench} pairs it with fault injection, and
\citet{liu2026physicianbench} applies that verdict-level idea to a clinical agent.
\citet{yagubyan2026consistent} does reach the calls, asking whether an agent selects
the same tools with the same arguments across identical invocations, but outside
medicine. Closest of all, \citet{mokssit2026fhiragenteval} rerun a clinical FHIR
sandbox three times per task and execute its writes against a resettable server, but
their run-to-run variance is over task success: the payloads and field values are
never compared. What none compares is the order itself across identical runs: its
payload, its destination, and the verdict returned for it.
Section~\ref{sec:metrics} defines four action-level quantities that do, and
Section~\ref{sec:results} measures them.

The same shift, from grading what an agent says to grading what it does, has been
argued for fairness \citep{morla2026agentfairbench}; reliability has not had the
equivalent treatment. Automated judges also move under rephrased prompts
\citep{bellibatlu2026judgesense}, though our grader is a fixed reference
implementation rather than a model, and adjacent work finds clinical agents acting
without noticing patient-identity errors \citep{klang2026identity} and benchmark
scores diverging from deployment behaviour
\citep{agrawal2025illusion,mehandru2024agents}.

\section{Environment}
\label{sec:environment}

\paragraph{MedAgentBench.}
MedAgentBench \citep{jiang2025medagentbench} provides 300
clinically derived tasks in ten families, a FHIR-compliant record server holding
records its authors describe as covering about one hundred patients, and a text
protocol in which the agent
issues \texttt{GET} and \texttt{POST} requests and ends with \texttt{FINISH}. Five
families can involve a write: one records an observation, one places a service
request, and three are conditional on a laboratory value or its age, one of those
three pairing a medication with a service request. In the conditional families
ordering nothing is sometimes correct, so a group there can pass without writing,
though inputs are fixed within a group and the right answer is the same across its
five runs. We use all five, since only there does an action exist to diverge.

A second version \citep{chen2026medagentbenchv2} revises the agent scaffold and
leaves the environment alone: a new system prompt, few-shot examples, memory, and
tools that let the agent act without constructing HTTP requests by hand, reaching
91\% success with GPT-4.1. Our measurements use the original scaffold, and
Section~\ref{sec:limitations} states what that does and does not imply. The record server ships as a container image; lacking a container runtime we
pulled the layers from the registry and ran the extracted HAPI FHIR server locally,
releasing the script and the store's resource counts (Appendix~\ref{app:impl}).

\paragraph{How the benchmark handles writes.}
The benchmark never forwards an agent's write to its record server, and says so: its
authors ``decide to only send GET requests to the environment'' because it ``takes
around 90 seconds to start'', and on a
\texttt{POST} they ``conduct a simple sanity check to make sure the payload data is
JSON-loadable, and indicate success of execution to the agent system''
\citep{jiang2025medagentbench}. The grader then scores from the transcript. No state
reset is needed between runs, which we confirmed: agents emitted 189
\texttt{ServiceRequest} payloads across the 1000 runs, and the store still held zero
afterwards. Two departures from the reference implementation remain: agent turns
keep their message roles, and \texttt{GET} responses are truncated at 6000
characters. Neither touches the task set, the prompt or the grader
(Appendix~\ref{app:impl}).

% BEGIN GENERATED MAIN TABLE
% Written by analysis/main_table.py -- do not edit by hand.
% Source files and run counts at generation time:
% data/grid_qwen4b_t005.jsonl -- 250 runs, 50 task groups
% data/grid_qwen4b_t07.jsonl -- 250 runs, 50 task groups
% data/grid_llama8b_t005.jsonl -- 250 runs, 50 task groups
% data/grid_llama8b_t07.jsonl -- 250 runs, 50 task groups
\begin{table*}[t]
\centering
\footnotesize
\caption{Divergence across five identical reruns of each task. Action quantities are computed over the task groups that emitted at least one order, since a task that never orders cannot diverge in its orders; trajectory and verdict quantities are over all groups, apart from the last row, which repeats verdict instability over the ordering groups; the mean pass rate is over the arm's 250 runs. \emph{The columns are not a ranking.} Each arm carries its own denominator, and we ran no cross-model test.}
\label{tab:main}
\begin{tabular}{lcccc}
\toprule
 & Qwen 0.05 & Qwen 0.7 & Llama 0.05 & Llama 0.7 \\
\midrule
Task groups & 50 & 50 & 50 & 50 \\
Groups emitting any order & 20 & 20 & 21 & 43 \\
Mean pass rate & 0.31 & 0.30 & 0.22 & 0.18 \\
\midrule
\multicolumn{5}{l}{\emph{Action level}} \\
Order-set divergence & 3/20 {\scriptsize [0.05, 0.36]} & 10/20 {\scriptsize [0.30, 0.70]} & 11/21 {\scriptsize [0.32, 0.72]} & 43/43 {\scriptsize [0.92, 1.00]} \\
\quad as pairwise disagreement & 0.08 & 0.29 & 0.33 & 0.78 \\
Omission split & 0/20 {\scriptsize [0.00, 0.16]} & 4/20 {\scriptsize [0.08, 0.42]} & 1/21 {\scriptsize [0.01, 0.23]} & 26/43 {\scriptsize [0.46, 0.74]} \\
Endpoint divergence & 3/20 {\scriptsize [0.05, 0.36]} & 8/20 {\scriptsize [0.22, 0.61]} & 8/21 {\scriptsize [0.21, 0.59]} & 12/43 {\scriptsize [0.17, 0.43]} \\
Clinical-content divergence & 0/20 {\scriptsize [0.00, 0.16]} & 0/20 {\scriptsize [0.00, 0.16]} & 0/21 {\scriptsize [0.00, 0.15]} & 28/43 {\scriptsize [0.50, 0.78]} \\
\midrule
\multicolumn{5}{l}{\emph{Trajectory and verdict}} \\
Trajectory divergence & 28/50 {\scriptsize [0.42, 0.69]} & 49/50 {\scriptsize [0.90, 1.00]} & 35/50 {\scriptsize [0.56, 0.81]} & 50/50 {\scriptsize [0.93, 1.00]} \\
Verdict instability & 5/50 {\scriptsize [0.04, 0.21]} & 12/50 {\scriptsize [0.14, 0.37]} & 11/50 {\scriptsize [0.13, 0.35]} & 21/50 {\scriptsize [0.29, 0.56]} \\
Verdict instability, ordering & 0/20 {\scriptsize [0.00, 0.16]} & 0/20 {\scriptsize [0.00, 0.16]} & 8/21 {\scriptsize [0.21, 0.59]} & 21/43 {\scriptsize [0.35, 0.63]} \\
\bottomrule
\end{tabular}
\end{table*}
% END GENERATED MAIN TABLE

\section{Protocol and metrics}
\label{sec:metrics}

The request is what stays comparable across runs, and comparing requests is the
whole of the protocol, which we call \emph{same-input rerun}. Figure~\ref{fig:protocol}
shows its shape: for a fixed task, model and decoding configuration the benchmark's
agent loop runs five times with every input identical, and the five runs are compared
at three layers. The task is its own control, so a difference cannot be attributed to
the case, the prompt or the environment, none of which changed.

Six quantities are reported.

A task that never orders cannot diverge in its orders, so the four action-level
quantities are computed only over groups that emit at least one order in at least
one run, the \emph{ordering groups}. The runs determine that set, so it differs by
arm: Llama orders in 21 groups at 0.05 and 43 under sampling. Every action-level
rate carries its own arm's denominator, and the paired tests below handle the
problem differently.

\paragraph{Order-set divergence.} For each task group we count the distinct
canonical order sets across its five runs, the canonical form being the destination
endpoint together with the whole payload, compared as a set. Nothing is discarded,
the clock and workflow fields included, and no group in any arm diverges on those
alone (Appendix~\ref{app:impl}). A run that emitted nothing contributes the empty
set, so an omitted order counts as a difference here, where endpoint and
clinical-content divergence exclude it. A group is divergent if that count exceeds
one. Section~\ref{sec:endpoint} shows why the endpoint is part of the action rather
than a detail of how it was written.

\paragraph{Omission split.} A group is split if at least one run emitted an order
and at least one emitted none; we report the proportion of ordering groups that are.
This is the sharpest of the six: the failure it names is silent, nothing in the
transcript announcing a missing order. In the conditional families a split can
equally be an order placed where none was called for; we do not separate the two.

\paragraph{Endpoint divergence.} For each run that emitted we take the set of
endpoints its orders reached, the endpoint being the final path segment of the
request URL, and a group diverges if two such runs disagree. Runs that emitted
nothing are excluded rather than entered as empty sets, so dropping an order cannot
by itself make a group divergent and this measure is not determined by the omission
split, though a group may do both and omission censors it, since a group left with
one emitting run can never be counted divergent (Appendix~\ref{app:impl}). It is
reported separately because a destination error and a payload error need different
fixes.

\paragraph{Trajectory first-divergence.} The index of the first agent turn at which
runs stop agreeing, reported as a distribution over diverging groups and, in
Table~\ref{tab:main}, as the proportion of groups that diverge at all. Query
parameters are sorted first, since \texttt{?patient=X\&code=Y} and
\texttt{?code=Y\&patient=X} are the same request; that correction changes at most
one group per arm (Appendix~\ref{app:impl}).

\paragraph{Clinical-content divergence.} Two runs may emit the same number of orders
to the same endpoint and still order different things. For every pair of runs that
both emitted we compare the fields carrying clinical content: the resource code and
the resource's own \texttt{text} element, the recorded value, the medication code
and its text, and the dose value, unit, rate and route. A group diverges if any of
those differs, and we report the proportion of ordering groups that do. Timestamps
and workflow status are excluded, since they record when a payload was written and
in what state rather than what was ordered, as are runs that emitted nothing, so the
omission split cannot account for this measure. The human-readable label a coding carries alongside its code is also
excluded, since it names the code rather than changing it. That exclusion is
load-bearing, and Appendix~\ref{app:impl} gives its cost: counting the label would
put this measure above zero in three arms rather than one. The comparison is
otherwise syntactic, so two clinically equivalent codings would count as divergent,
which can only inflate this rate.

\paragraph{Verdict instability.} A group is unstable if the benchmark's own
pass/fail judgement is not identical across its five runs; we report the proportion
over all groups and, separately, over ordering groups, so it can be compared with
the action-level rates on a matched denominator. This is what a leaderboard reports,
included so it can be measured against the other five rather than assumed to stand
for them. It is also, with trajectory, a quantity a group can move without ordering
anything, by changing the answer it reports.

\paragraph{A rate that survives a change of budget.} Each of the four action-level
quantities asks whether a group diverged at least once in five runs, and that
question is monotone in the rerun budget: run the same agent ten times and the rate
can only rise, so a figure measured at five runs cannot be set beside one measured
at three. We therefore report order-set divergence a second way, as \emph{pairwise
disagreement}: the share of run pairs within a group whose order sets differ,
averaged over the ordering groups. For exchangeable runs it estimates the same
quantity at any budget of two or more, though which groups enter the average still
depends on that budget. The at-least-once form says whether a task ever went wrong;
the pairwise form is the one another study can be read against.

Proportions carry Wilson 95\% intervals, which remain defined at these denominators
and at zero or one. Between the two temperatures of a model we use an exact paired
McNemar test and state the discordant counts, since at these sizes a
non-significant result is not evidence of similarity. The tests run over all 50 task
groups rather than the ordering groups, because a group that orders in one arm and
not the other is informative about the temperature effect and dropping it would
condition on an outcome. A group that never orders in either arm is concordant by
construction on order-set divergence, though not on trajectory or verdict, where it
can still move (Section~\ref{sec:results}).
Appendix~\ref{app:mcnemar} gives each test's discordant counts and denominator. The
corrected family is one quantity per comparison layer per model, six in all; the
other three action rows decompose order-set divergence rather than standing beside
it, so they are read descriptively and carry no corrected $p$. Task groups cluster into
families rather than being independent, so we bootstrap over the families that
contain ordering groups: five for Llama under sampling, three at its near-greedy
setting, and two in each Qwen arm. With two or three clusters the interval is little
more than the range of the family rates, and should be read as a robustness check
rather than an estimate.

\section{Experimental setup}
\label{sec:setup}

The grid is small and fully crossed: two models, two decoding temperatures, and the
same 50 tasks throughout, so that a difference between the two temperatures of a
model cannot come from the task sample. Differences between models are a separate
matter, and Table~\ref{tab:main} says why we do not read them.

\paragraph{Models.} Two open-weight instruction-tuned models, Qwen3-4B and
Llama-3.1-8B, served locally and quantised to four bits
(Appendix~\ref{app:impl}). The panel is restricted to models an institution could
run on its own hardware, the deployment mode available where patient data cannot
be sent to a hosted API. It is not a frontier panel (Section~\ref{sec:limitations}).

% HAN 2026-09-09 (verbatim): The task-selection procedure is underspecified. The
% experimental setup states that 10 tasks were selected from each task family, but
% it does not explain how these tasks were chosen, for example, by random sampling,
% by taking the first 10 tasks, or according to specific inclusion or exclusion
% criteria.
% HAN-STATUS: done. Checked against harness/medagent_runner.py and
% data/write_tasks_50.txt: it is the first ten identifiers of each family, a
% deterministic prefix of thirty. Stated here and recorded as a limitation.
\paragraph{Decoding.} MedAgentBench evaluates at temperature 0 and scores a single
attempt per task, stating that it adopts \mbox{pass@1} because a single incorrect
action can have significant consequences \citep{jiang2025medagentbench}. We run each
model at 0.05, a near-greedy setting, and at 0.7 with nucleus sampling. Neither is
the benchmark's own value, and the choice is deliberate: the noise floor below shows
greedy decoding on this stack to be exactly reproducible, so a study run at 0 would
measure nothing here.

That reproducibility belongs to the stack, and it does not survive a shared server.
We serve one request at a time on one local GPU; a deployment serving many clinicians
batches them, and batch composition changes which floating-point reductions run
\citep{atil2024nondeterminism,yuan2025numerical}. Reissuing the noise floor with
other traffic in flight breaks determinism at temperature zero in both models: four
of twelve greedy cells return more than one completion of 20, where every matched
serial cell returned one (Appendix~\ref{app:impl}). What moves is completion text
rather than an order, but the benchmark's own setting guarantees no run-to-run
identity where a clinical agent would run, and a fixed seed does not close the gap,
since seeding makes five runs one draw repeated.

\paragraph{Tasks.} Each family holds 30 tasks; we take the first ten of each in the
benchmark's own identifier order, a deterministic prefix rather than a random
sample, and run each five times, giving 250 runs per arm and 1000 in total. The
round limit of eight is the reference implementation's, and the identifiers are
released with the harness.

\paragraph{Controls.} Nothing is trained here, so there is no baseline and no data
split; the controls are on the instrument instead, the record-server check of
Section~\ref{sec:endpoint} and a noise floor. Before attributing anything to the
agent we measured what the serving stack contributes when nothing should vary:
twenty identical calls per cell over eighteen cells (Appendix~\ref{app:impl}).
Greedy decoding here is exactly reproducible, seeded or not, so downstream variation
belongs to the decoding configuration and not to infrastructure. This replicates
\citet{atil2024nondeterminism}, who report the same determinism running a model on
their own hardware without optimisations.

\section{Is endpoint divergence cosmetic?}
\label{sec:endpoint}

Endpoint divergence counts as an action difference only if the two destinations
behave differently, and it is reasonable to ask whether posting to \texttt{/fhir/}
with \texttt{resourceType} in the body is equivalent to posting to
\texttt{/fhir/ServiceRequest}, since some servers resolve the type from the payload.
The question is decidable against the same HAPI FHIR server the benchmark ships, so
we decided it. An identical \texttt{ServiceRequest} payload sent to the typed
endpoint returned \texttt{201 Created} and the resource was created, which we then
deleted; sent to the collection root it returned \texttt{400 Bad Request} with
\texttt{HAPI-0450: Failed to parse request body as JSON resource}.

The two are therefore not interchangeable: one places the order and the other does
not. The benchmark's paper tells its human readers that writes are not executed, but
the agent reads only the runtime reply, which reports success for any payload the
sanity check parses, whatever the destination. The gap is not the absent write,
which is disclosed and deliberate, but that the surface the agent acts on carries no
trace of it.

\section{Results}
\label{sec:results}

\begin{figure}[t]
\centering
\includegraphics[width=0.96\columnwidth]{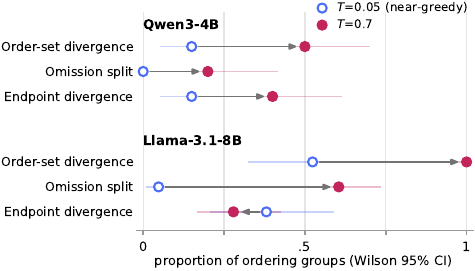}
\caption{Three of the four action-level rows of Table~\ref{tab:main};
clinical-content divergence is zero in every arm but one. Drawn as the movement from
$T{=}0.05$ (open mark) to $T{=}0.7$ (solid), with Wilson intervals. Each arm's
denominator is its own ordering groups: 20 and 20 for Qwen, 21 and 43 for Llama.
Only Llama's grows under sampling, which is why its endpoint arrow points left as
the count rises from 8 to 12.}
\label{fig:divergence}
\end{figure}

Order-set divergence and the omission split rise with temperature in both models,
and clinical-content divergence appears in one arm only. Endpoint divergence is the
exception: for Llama it falls as a rate because the base grows faster than the
count. Figure~\ref{fig:divergence} plots three of those rows as the move from
near-greedy to sampling.

A group whose orders differ has always taken a different path to them, so trajectory
divergence is the weaker condition: over the ordering groups it exceeds order-set
divergence in both Qwen arms and coincides with it in both Llama arms
(Appendix~\ref{app:impl}). Divergence is immediate rather than
cumulative, the median first-divergence index being zero in three arms of four and
never exceeding four (Appendix~\ref{app:impl}).

\paragraph{What the tests do and do not establish.}
Paired exact McNemar tests over all 50 task groups find the rise with temperature
significant for both models on trajectory, order-set and verdict divergence, with 96
of the 100 discordant pairs running toward more divergence and Holm leaving all six
significant at 0.047. Two of the six need care. Llama's ordering base grows from 21 groups to
43, so most of the 33 groups divergent only at 0.7 began ordering at all rather than
became inconsistent; restricted to the 20 that order at both temperatures the effect
holds, 10 discordant to none the other way ($p=0.002$). And all seven discordant
groups in Qwen's verdict test never order in either arm, so what moves there is the
answer the agent reports, not an order. The tests also treat task groups as
independent, which Section~\ref{sec:metrics} says they are not. Taking each family as one signed
observation instead, no family in any test moves back, but only one to four move at
all, and an exact sign test on that many cannot fall below $p=0.125$. The direction is consistent at both units; the significance is not,
and we rest nothing on it. The pairwise row of Table~\ref{tab:main} carries the
same rise in the form that does not grow with the rerun budget, 0.08 to 0.29 for
Qwen and 0.33 to 0.78 for Llama.

\paragraph{A number belongs to a model and a configuration.}
Llama-3.1-8B shows more order-set divergence and more omission than Qwen3-4B at both
temperatures, which we do not read as a model difference: the rates overlap once
clustered, and we ran no cross-model test (Appendix~\ref{app:impl}).

\paragraph{Two mechanisms are mechanical, and a third is not.}
Orders are misrouted, reaching the typed endpoint on some runs and the collection
root on others, which Section~\ref{sec:endpoint} shows is the difference between a
created order and a rejected one. Orders are also lost: in three service-request
groups under Qwen sampling the referral is placed on four runs of five, while on the
fifth the payload fails to parse and the retry repeats the same malformed
construction, a protocol failure rather than a change of mind. Nor is the loss
the agent running out of turns: of 72 non-emitting runs in Llama's 26 split groups,
61 ended with the agent declaring itself finished, 7 on a malformed action and 4 at
the round limit.

The third is not. In the Llama sampling arm 28 of the 43 ordering groups post a
different value across identical runs: 23 differ on the resource code, 18 on a
recorded value, 5 on the medication and 3 on the dose, categories that overlap. The count covers
only runs that emitted, so omission cannot explain it, and only fields carrying
clinical content. No other arm of the grid shows any, and
Appendix~\ref{app:impl} gives two worked instances.

This measure is the one our design can least well attribute, appearing in one arm
only, where four-bit quantisation and sampling compound. The eight-bit Qwen rerun
complicates a simple quantisation account, since content divergence appears there at
eight bits and not at four, though in Qwen rather than the Llama arm that shows it
here. What we stand behind is that
divergence reaches the content of an order, not only its envelope.

\paragraph{Orders move without the verdict moving.}
Restricting both to the ordering groups, what matters is the gap: groups whose
orders differ across identical runs while the verdict holds. In the Llama sampling
arm 22 of the 43 divergent groups are of that kind, spanning all five write
families; at the near-greedy setting it is 3 of 11, and for Qwen under sampling 10
of 10. That verdict is almost always a failure: of the 38 such groups across the
four arms, 36 fail on all five runs, and the 2 that pass differ only in the label
attached to an identical procedure code. What the score conceals is therefore not
success but that the agent failed differently each time; none of the 251 passing
runs across the grid posted to a destination the server would refuse.

The converse, that no verdict moves while the orders hold, is a property of the
grader: in three of the five families it decides from the emitted request alone, so
a verdict cannot move there unless the order set does, and 25 of the 29
ordering-group verdict flips fall there. We draw no conclusion in that direction.

A natural objection is that this only measures agents too weak to do the task. Llama
at the near-greedy setting diverges in all 10 ordering groups it ever passes against
1 of the 11 it never passes, but eight of those ten pass only on some runs, and in a
family whose grader reads only the request a change of verdict and a change of order
are one event rather than two pieces of evidence. Two groups pass on all five runs
and diverge anyway, but in the label rather than the code, so
what reaches a task the agent completes every time is instability of form. No group
in this study both passes every run and changes what it ordered. The objection
therefore stands to this extent: we show that agents fail inconsistently, and
showing that they succeed inconsistently would need a stronger panel.

Order-set divergence is not independent of the rows below it in
Table~\ref{tab:main}: a group counts as divergent if it omitted the order, sent it
elsewhere or changed the payload, so the 43 of 43 is their union rather than a third
finding, and Appendix~\ref{app:impl} gives the decomposition.

\paragraph{The effect is not spread evenly, and thins under clustering.}
Table~\ref{tab:family} breaks the ordering groups down by family. Qwen's entire
action-level temperature effect sits in one family, its service requests going from
3 of 10 divergent to 10 of 10, while its observation tasks never diverge and its
other three never order at all; Llama under sampling diverges in every family that
orders, so concentration belongs to the quieter arms rather than to the measure.
Resampling families rather than groups then widens the intervals sharply: two arms
collapse to $[0.00, 1.00]$, Qwen at 0.7 and Llama at 0.05, Qwen at 0.05 reaches
$[0.00, 0.30]$, and only the Llama sampling arm holds, at $[1.00, 1.00]$, every one
of its five ordering families diverging. The study establishes that action divergence
occurs and can pass unrecorded by the score, not that any rate generalises: a
benchmark can report a stable number while the behaviour under it moves.

% HAN 2026-09-09 (verbatim): Gaps and solutions are too far apart. they better stay
% adjacent. but it requires article structures reordered, you can try to solve this
% problem.
% HAN-STATUS: open. The distance is shorter, not gone. The three changes now appear
% in the introduction immediately after the gap, the abstract closes on them, and
% the background is shorter, so a reader meets the remedy on page 2 rather than
% page 8. The sections themselves have NOT been reordered; that is a structural
% rewrite and it is the one item of his still outstanding.
\section{What this suggests for benchmark design}
\label{sec:recommendations}

Three changes would make clinical agent benchmarks report what readers take them to
report. None is a research programme; each can be adopted on its own.

\paragraph{R1. Rerun, and say so.} MedAgentBench states its protocol: a single
attempt per task, reasoning that one wrong action can matter clinically
\citep{jiang2025medagentbench}. That argues for reruns, since one attempt cannot
show whether the wrong action recurs, and guidance for single-turn evaluation
already suggests at least two \citep{alvarado2025repetitions}.

\paragraph{R2. Report an action-level stability measure.} No clinical agent
benchmark we read in detail reports one: \citet{jiang2025medagentbench} scores a
single attempt, while
\citet{liu2026physicianbench} and \citet{mokssit2026fhiragenteval} rerun but score
the outcome rather than the order. Order-set divergence and the omission split can be computed from transcripts
a benchmark already stores, and we would report the first in its pairwise form, so
that studies bought at different rerun budgets stay comparable.

\paragraph{R3. Let the environment answer as the real system would.} A request the FHIR
server would reject is reported to the agent as executed, removing the signal it would need to
recover. The stated reason for not executing writes is the cost of re-initialising
the environment, which covers an accepted write; a refused one leaves nothing to
reset, and the refusal would make misrouting visible to the agent and the score. One clinical sandbox already executes writes against a
resettable server \citep{mokssit2026fhiragenteval}.

\section{Limitations}
\label{sec:limitations}

\paragraph{Emitted, not executed.} We do not claim clinical agents are unsafe to
deploy: the benchmark forwards no writes, so every statement here concerns the order
issued, not the order received.

\paragraph{Scaffold, model panel and scope.} The benchmark's second version removes
hand-constructed HTTP requests as error-prone \citep{chen2026medagentbenchv2}, the
mechanism behind many failures here, so our rates describe the original scaffold,
though a higher success rate need not imply action stability. The panel is two
open-weight models below ten billion
parameters, quantised to four bits, on one benchmark and one serving stack, and a
prefix of each family rather than a random sample, which under-represents harder tasks if difficulty rises with
identifier order. An eight-bit rerun of the Qwen sampling arm separates part of
this: the ten groups carrying its divergence diverge at eight bits too, and only the
mode of failure changes (Appendix~\ref{app:impl}). A frontier model might show none
of it, though reaching one means a hosted API, while this panel is the deployment
where records stay inside the institution. The protocol transfers; the rates do not.

% HAN 2026-09-09 (verbatim): Gap, question introduced at beginning, but no
% conclusive idea at the end.
% HAN-STATUS: done. The conclusion closed on what had been released, which is the
% same fault he named in the abstract. It now closes on what a benchmark has to do
% differently, so the proposal is stated four times: abstract, introduction,
% Section 8, conclusion.
% The conclusion runs a few lines onto the reference page. Every relocatable
% passage has already gone to an appendix, so the page is lengthened instead of
% the argument shortened. Remove if the layout changes.
% Two lines, not six: the six-line version pushed the last page into the bottom
% margin, which a reviewer reads as a formatting violation. Everything relocatable
% has already gone to an appendix.
% The body runs a few lines onto the reference page without this. Everything
% relocatable has already gone to an appendix; four lines keeps the page bottom
% within a couple of points of its normal position. Remove if the layout changes.
% No page limit on arXiv, so the last page is not stretched to hold the body to
% eight pages. The submission build keeps \enlargethispage; this one does not need it.
\section{Conclusion}

One run says whether the agent was right, not whether it is consistent. A benchmark
for clinical actions must rerun, measure the order as well as the verdict, and
answer as the record system would.
%
% The body must end on page 8. With \raggedbottom only text near the end moves the
% boundary, so trim here or in Sections 8 and 9, never earlier.

\bibliography{ref}

% References precede the appendix, which starts a fresh page and runs in one
% column so the table sits under its own heading instead of floating past it.
\clearpage
\appendix
\onecolumn

\section{Paired-test discordance tables}
\label{app:mcnemar}

% BEGIN GENERATED MCNEMAR TABLE
% Written by analysis/mcnemar_table.py -- do not edit by hand.
\begin{table}[h]
\centering
\footnotesize
\caption{The six paired tests, with the discordance behind each. $b$ counts groups showing the measure only at $T{=}0.7$ and $c$ only at $T{=}0.05$; the exact two-sided McNemar $p$ uses $b$ and $c$ alone. All six run over the 50 task groups. A group that never orders in either arm is concordant by construction on order-set divergence, but not on the other two.}
\label{tab:mcnemar}
\begin{tabular}{llccccc}
\toprule
Model & Measure & $n$ & $b$ & $c$ & disc. & $p$ \\
\midrule
Qwen3-4B & Trajectory divergence & 50 & 21 & 0 & 21 & $<$0.0001 \\
Qwen3-4B & Order-set divergence & 50 & 7 & 0 & 7 & 0.0156 \\
Qwen3-4B & Verdict instability & 50 & 7 & 0 & 7 & 0.0156 \\
\midrule
Llama-3.1-8B & Trajectory divergence & 50 & 15 & 0 & 15 & $<$0.0001 \\
Llama-3.1-8B & Order-set divergence & 50 & 33 & 1 & 34 & $<$0.0001 \\
Llama-3.1-8B & Verdict instability & 50 & 13 & 3 & 16 & 0.0213 \\
\bottomrule
\end{tabular}
\end{table}
% END GENERATED MCNEMAR TABLE

\section{Implementation details}

\label{app:impl}

\paragraph{Serving stack.} Both models are served through Ollama at four-bit
quantisation (\texttt{q4\_K\_M}) on a single consumer GPU with an 8192-token
context window. The exact tags are \texttt{qwen3:4b-instruct-2507-q4\_K\_M} and
\texttt{llama3.1:8b-instruct-q4\_K\_M}; the quantisation suffix is part of the tag,
and pulling either name without it gives a different build. Sampling runs use
nucleus $p=0.9$. The round limit of eight turns is that of the reference
implementation.

\paragraph{Why the harness departs from the reference implementation.} Agent turns
are sent as multi-turn messages with roles preserved. An earlier version flattened
the dialogue into a single prompt, under which nearly every task failed on output
formatting before reaching the reasoning it was meant to test, which would have
measured the harness rather than the agent. \texttt{GET} responses are truncated to
6000 characters because unbounded FHIR bundles exhaust the context window of a
locally served model; truncation fired on 18 and 11 of 250 runs in the Qwen arms
and 26 and 10 of 250 in the Llama arms, and is recorded in every run record.
Truncation bounds rather than explains the omissions of
Section~\ref{sec:results}: of the 72 non-emitting runs in the 26 omission-split
groups under Llama sampling, 7 saw a truncated response, against 2 of the 58 runs in
those groups that did emit. Truncation is thus about three times as frequent where
the order went missing, and could at most account for 7 of the 72; the remaining 65
non-emitting runs saw the response in full. Those seven are among the runs that
finished, not the seven that ended on a malformed action.

\paragraph{What order-set divergence is made of.} In the Llama sampling arm 26 of
the 43 divergent groups diverge through omission and 12 through misrouting,
overlapping on 5, so those two cover 33; the remaining 10 differ in payload alone,
every one on an observation code or a recorded value. The five in the overlap drop
the order on some run and split the destination across the runs that remain.

\paragraph{The coding label.} Clinical-content divergence compares the code a
resource carries, not the display string beside it: two runs that both post SNOMED
306181000000106, one labelling it \emph{Orthopaedic surgical referral} and the other
\emph{Orthopedic surgery referral}, have ordered the same procedure. Counting the
label as content would raise the measure from none to 10 of 21 in the Llama
near-greedy arm, from 28 to 33 of 43 under sampling, and from none to 1 of 20 for
Qwen at 0.7. The finding that content divergence is confined to one arm is therefore
a finding about what was ordered, not about how it was labelled, and the two groups
of Section~\ref{sec:results} that pass every run while their orders differ are label
differences of exactly this kind.

\paragraph{The eight-bit arm.} To separate the agent's contribution from the
four-bit quantisation, the Qwen sampling arm was rerun as
\texttt{qwen3:4b-instruct-2507-q8\_0} with everything else held fixed: the same 50
task groups, the same temperature and nucleus setting, the same 8192-token context
and the same five reruns. Its records are released alongside the grid. Across all 50
task groups the eight-bit arm orders in 22 against the four-bit arm's 20 and diverges
in 12 of those against 10, while endpoint divergence falls from 8 of 20 to none of
22 and omission rises from 4 of 20 to 12 of 22. Over the ten groups that carry the
four-bit arm's divergence, all ten diverge at both precisions, endpoint divergence
falls from 8 of 10 to none, omission rises from 4 of 10 to all ten, and
clinical-content divergence appears in 3 of 10 where four bits shows none. Higher
precision therefore decides which failure occurs rather than whether one occurs.

\paragraph{Endpoint divergence and single-emitter groups.} Endpoint divergence
compares the endpoint sets of runs that emitted, so a group left with one emitting
run cannot be counted divergent. Seven of the 43 ordering groups in the Llama
sampling arm are in that position, one of 21 at the near-greedy setting, and none
in either Qwen arm. Table~\ref{tab:main} keeps the ordering groups as the
denominator so that all four action rows are read against the same population; over
the groups that could show it, the Llama sampling rates are 12 of 36 for endpoint
divergence and 28 of 36 for clinical content, and 8 of 20 and 0 of 20 at the
near-greedy setting.

\paragraph{Noise floor cells.} Twenty identical calls were issued per cell across
two models, three decoding configurations and three clinical prompts, giving 18
cells. The prompts ask for a treatment choice, for the most urgent action in a
deteriorating patient, and for one laboratory value read out of a note; the first
two are referred to below as the treatment and triage prompts. All twelve greedy cells returned twenty identical completions, with and
without a fixed seed. Under sampling the two open-ended prompts produced between 14
and 20 distinct completions of 20, while the prompt asking only for a numeric value
extracted from a note produced 1 for one model and 3 for the other: sampling does
not perturb an output that has essentially one form.

\paragraph{The noise floor under batching.} The cells above serve one request at a
time. A deployment does not, so we reissued them with other traffic in flight: each
trial sends the target request together with one to three filler requests of
differing length, so the batch it joins differs from trial to trial, and every other
input is held fixed. The server is configured to decode four sequences at once, and
four concurrent requests complete in 8.3 seconds against 21.9 if run one after
another, so the requests share batches rather than queueing. Over 240 such trials,
four of the twelve greedy cells stop being reproducible, two in each model, against
one distinct completion in all twelve matched serial cells. Which cell moves depends
on the model: the 4B model diverges on the open-ended triage prompt, three
completions of 20 both with the seed fixed and without it, while the 8B model
diverges on the treatment prompt, two of 20 seeded and three unseeded. The prompt
asking only for a number extracted from a note never moves in either. Divergence
here is textual: the three triage completions name the same action and differ in
wording and in the shock they attribute it to. This is a property of the serving
stack rather than of the agent, and it is why a benchmark evaluating at temperature
zero cannot infer from that setting alone that its runs repeat in deployment.

\paragraph{Record store.} The store holds 695 \texttt{Patient} resources, of which
the 300 benchmark tasks name 98, so most of the cohort is background a query has to
search, plus 563{,}426 observations and 124{,}969 procedures.

\paragraph{Clinical-content divergence, two instances.} In one observation task
under Llama sampling, the same blood pressure is recorded under LOINC 8480-6 with
value \texttt{`118/77'} on one run, under 8462-4 with value 118 on a second, and
under 8462-4 with value 77 on a third. In a magnesium task, one run posts a
\texttt{MedicationRequest} for one gram and another posts an \texttt{Observation}.
Writing the systolic value under a diastolic code is not a difference of clinical
judgement, though what produces it is what Section~\ref{sec:results} says we cannot
settle.

\paragraph{The two models side by side.} Llama diverges in 11 of its 21 ordering
groups at the near-greedy setting against Qwen's 3 of 20, but that Llama rate
carries a family-level interval of $[0.00, 1.00]$ and overlaps almost entirely with
Qwen's at 0.7, 10 of 20. We ran no cross-model test and read no model difference
from the pair.

\paragraph{Trajectory against orders.} Restricted to each arm's ordering groups,
Qwen diverges in trajectory in 8 of 20 groups against 3 in their orders at 0.05, and
19 against 10 under sampling, while in both Llama arms the two coincide exactly, 11
of 21 and 43 of 43.

\paragraph{First-divergence index.} Of the groups that diverge at all, 12 of 28,
40 of 49, 27 of 35 and 50 of 50 part company on the very first agent turn, in the
column order of Table~\ref{tab:main}. The median index is 1, 0, 0 and 0 in that same
order, the largest anywhere being four, in the Qwen near-greedy arm. Runs separate
immediately far more often than they drift apart.

\paragraph{Clock and workflow fields.} Order-set divergence compares whole payloads,
so a differing \texttt{authoredOn}, \texttt{effectiveDateTime} or \texttt{status}
would count. In practice the benchmark hands the agent a fixed clock: those fields
vary within 11 of the 43 divergent groups in the Llama sampling arm and in none
elsewhere, and no group in any arm is divergent on them alone, so the reported
counts do not rest on a timestamp.

\paragraph{Query-parameter normalisation.} Across the four arms, raw comparison
finds 29, 49, 36 and 50 divergent groups of 50, against 28, 49, 35 and 50 after
sorting query parameters, so at most one group per arm diverges only in the order
of its query string.

\section{Divergence by task family}

The pooled rates in Table~\ref{tab:main} average over five task families that behave
very differently: F3 records an observation and F8 places a service request, both
unconditionally, while F5 orders magnesium replacement if the last level is recent
and low, F9 orders potassium and a follow-up draw if the last level is low, and F10
orders a repeat test if the last result is over a year old. In the latter three, a
run that orders nothing can be correct, and none of the three ever orders in either
Qwen arm: F9's pass rates of 0.46 and 0.40 there are runs that correctly withheld
the order. This breakdown is what that averaging hides, and it is
the evidence for the claim in Section~\ref{sec:results} that the pooled figures
summarise this task mixture rather than a general rate.

% BEGIN GENERATED FAMILY TABLE
% Written by analysis/family_table.py -- do not edit by hand.
\begin{table}[h]
\centering
\footnotesize
\caption{Divergent order sets out of ordering groups, with each family's mean pass rate in parentheses. Ten task groups per family per arm; a dash marks a family that never ordered in that arm.}
\label{tab:family}
\begin{tabular}{lcccc}
\toprule
Family & Qwen .05 & Qwen .7 & Llama .05 & Llama .7 \\
\midrule
F3 & 0/10 (1.00) & 0/10 (1.00) & 0/10 (0.00) & 10/10 (0.28) \\
F5 & -- (0.08) & -- (0.10) & 1/1 (0.50) & 8/8 (0.18) \\
F8 & 3/10 (0.00) & 10/10 (0.00) & 10/10 (0.58) & 10/10 (0.44) \\
F9 & -- (0.46) & -- (0.40) & -- (0.00) & 7/7 (0.00) \\
F10 & -- (0.00) & -- (0.00) & -- (0.00) & 8/8 (0.00) \\
\bottomrule
\end{tabular}
\end{table}
% END GENERATED FAMILY TABLE

\end{document}